\documentclass[letterpaper, 10 pt, journal, twoside]{ieeetran}

\IEEEoverridecommandlockouts
\usepackage{cite}
\usepackage{amsmath,amssymb,amsfonts}
\usepackage{algorithmic}
\usepackage{graphicx}
\usepackage{textcomp}
\usepackage{xcolor}
\def\BibTeX{{\rm B\kern-.05em{\sc i\kern-.025em b}\kern-.08em
    T\kern-.1667em\lower.7ex\hbox{E}\kern-.125emX}}

\usepackage{amsmath} 
\usepackage{amssymb}
\usepackage{amsfonts}
\usepackage{amstext}
\DeclareMathOperator*{\argmax}{arg\,max}
\DeclareMathOperator*{\argmin}{arg\,min}

\usepackage{pdfcomment}

\usepackage{mathtools}

\usepackage{graphicx}

\usepackage{booktabs}
\usepackage{multirow}

\usepackage{wrapfig}

\newcommand{\method}{\textit{REALM }}

\definecolor{Burgundy}{RGB}{144,0,32}

\usepackage{fancyhdr}
\fancypagestyle{plain}{
  \fancyhf{}
  \fancyhead[C]{Conference on \LaTeX}     
  \fancyfoot[L]{This is a notice}

}
\usepackage{eso-pic}
\usepackage{hyperref}

\begin{document}

\AddToShipoutPictureBG*{%
  \AtPageLowerLeft{%
    \setlength\unitlength{1in}%
    \hspace*{\dimexpr0.5\paperwidth\relax}
    \makebox(0,0.75)[c]{\parbox{0.7\paperwidth}{\raggedright\footnotesize© 2025 IEEE.  Personal use of this material is permitted. Permission from IEEE must be obtained for all other uses, in any current or future media, including reprinting/republishing this material for advertising or promotional purposes, creating new collective works, for resale or redistribution to servers or lists, or reuse of any copyrighted component of this work in other works.}}%
}}

\title{REALM: Real-Time Estimates of Assistance for \\Learned Models in Human-Robot Interaction}


\markboth{IEEE Robotics and Automation Letters. Preprint Version. Accepted April, 2025}
{Hagenow \MakeLowercase{\textit{et al.}}: REALM: Real-Time Estimates of Assistance for Learned Models in Human-Robot Interaction} 

\author{Michael Hagenow$^{1}$ and Julie A. Shah$^{1}$
\thanks{Manuscript received: December, 6, 2024; Revised February, 26, 2025; Accepted March, 30, 2025.}
\thanks{This paper was recommended for publication by Editor Angelika Peer upon evaluation of the Associate Editor and Reviewers' comments. This work was supported by the MIT Postdoctoral Fellowship Program for Engineering Excellence (PFPFEE). The authors also acknowledge the MIT SuperCloud and Lincoln Laboratory Supercomputing Center for providing HPC resources that have contributed to the reported research results.}
\thanks{$^{1}$Michael Hagenow and Julie Shah are with the Computer Science and Artificial Intelligence Lab (CSAIL), Massachusetts Institute of Technology, Cambridge, MA 02139, USA.
        {\tt\footnotesize \{hagenow,julie\_a\_shah\}@csail.mit.edu}}%
\thanks{Digital Object Identifier (DOI): \href{https://doi.org/10.1109/LRA.2025.3560862}{10.1109/LRA.2025.3560862}.}}%

\maketitle

\begin{abstract}
There are a variety of mechanisms (i.e., input types) for real-time human interaction that can facilitate effective human-robot teaming. For example, previous works have shown how teleoperation, corrective, and discrete (i.e., preference over a small number of choices) input can enable robots to complete complex tasks. However, few previous works have looked at combining different methods, and in particular, opportunities for a robot to estimate and elicit the most effective form of assistance given its understanding of a task. In this paper, we propose a method for estimating the value of different human assistance mechanisms based on the action uncertainty of a robot policy. Our key idea is to construct mathematical expressions for the expected post-interaction differential entropy (i.e., uncertainty) of a stochastic robot policy to compare the expected value of different interactions. As each type of human input imposes a different requirement for human involvement, we demonstrate how differential entropy estimates can be combined with a likelihood penalization approach to effectively balance feedback informational needs with the level of required input. We demonstrate evidence of how our approach interfaces with emergent learning models (e.g., a diffusion model) to produce accurate assistance value estimates through both simulation and a robot user study. Our user study results indicate that the proposed approach can enable task completion with minimal human feedback for uncertain robot behaviors.
\end{abstract}

\section{Introduction}

\IEEEPARstart{F}{or} complex and critical tasks, it is beneficial to maintain a skilled human operator in the loop who can ensure appropriate task outcomes. Within the broad umbrella of human-in-the-loop (HIL) methods, there are many different levels of automation and corresponding mechanisms of human input; including traded teleoperation (i.e., alternating periods of teleoperation and autonomy), control of robot subspaces (e.g., the human controls only rotation or position), and discrete input. However, few works have explored methods where robots elicit different levels of human feedback in real-time during task execution. Combining mechanisms offers an opportunity for robots to select their feedback requests in a way that balances the simplicity of human input with the informational needs of the robot. This work presents a method that mediates human assistance by estimating and selecting the most appropriate mechanism for real-time human feedback.


\begin{figure}[t]
\centering
\pdftooltip{\includegraphics[width=\columnwidth]{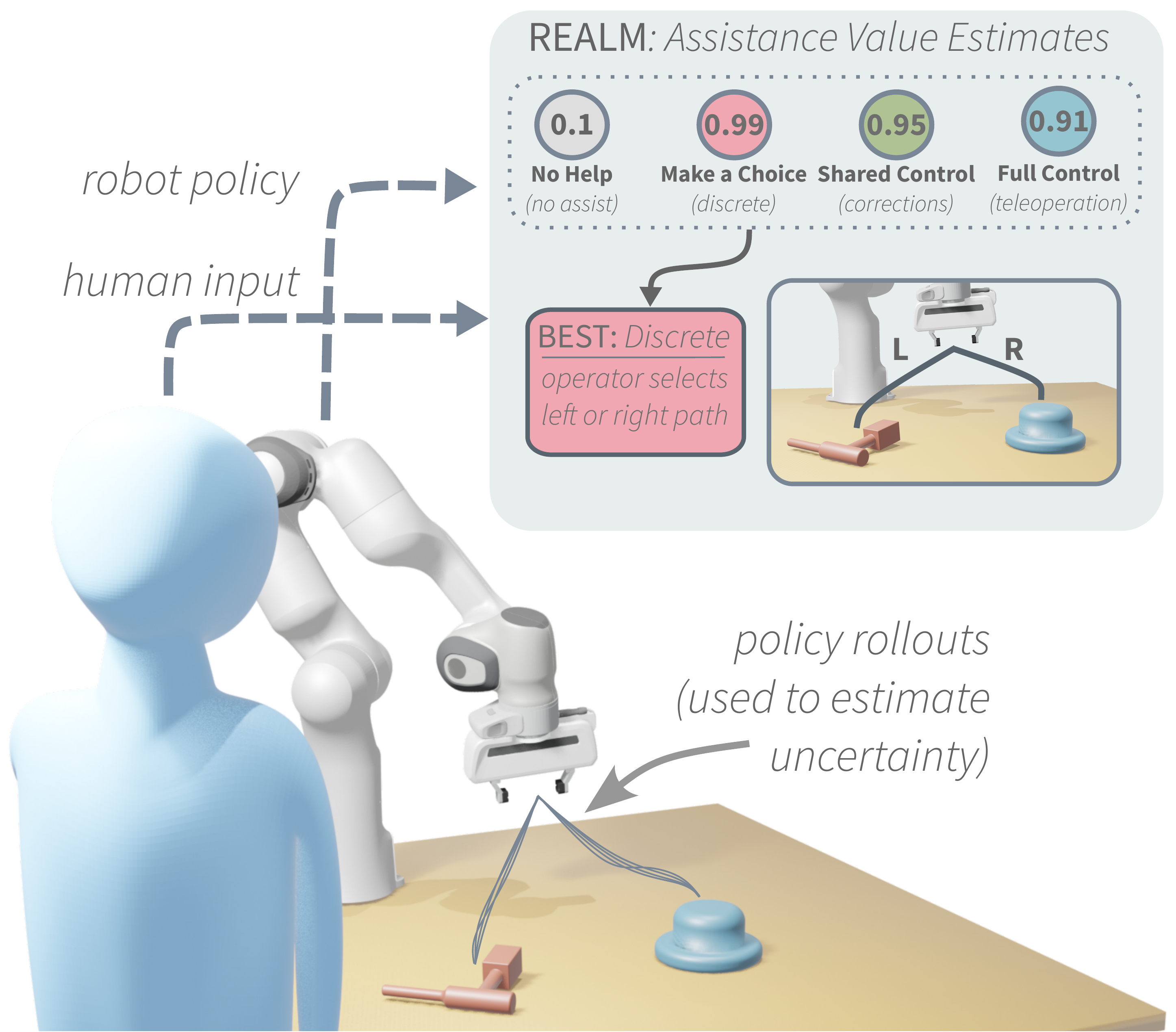}}{A human oversees a robot that is choosing whether to pick up a drill or a sanding tool. Sampled trajectories toward both objects indicate the robot is uncertain which object to pick up. These samples feed into a flow diagram that compares illustrative likelihoods for different human assistance types (0.1 for no help, 0.99 for making a discrete choice, 0.95 for corrective shared control, and 0.91 for teleoperation). The highest likelihood is selected and the human is asked to select the left or right path.}
\vspace{-15pt}
\caption{\textit{REALM} is a method to estimate the value of different types of human assistance during periods of robot uncertainty. Our method uses rollouts from a diffusion policy and an entropy-based formulation to assess the value of different human interactions. In this example, the policy rollouts (and corresponding entropy metrics) indicate a discrete intervention from the human is the most effective way to resolve the uncertainty.}
\label{fig:teaser}
\vspace{-15pt}
\end{figure}

Recent diffusion-based robot policies \cite{chi2023diffusion} are gaining traction in human-robot interaction \cite{yoneda2023noise,ng2023diffusion} and have made it possible to better capture the probability distribution of robot actions, which can serve as a measure of the robot's uncertainty. The pattern of this uncertainty can further serve as an indicator of the required granularity of human assistance. For example, if the uncertainty suggests there are two clear paths a robot might take (e.g., as shown in Figure \ref{fig:teaser}), a simple discrete choice from a human may be sufficient for the robot to proceed. If instead there are an uncountable number of paths, it may be most efficient for the person to temporarily teleoperate the robot. Our approach operationalizes this intuition by constructing entropy metrics that estimate the remaining uncertainty after candidate human interventions.





In this paper, we describe a method, \method  (\textit{Realtime Estimates of Assistance for Learned Models}), that leverages uncertainty in a stochastic robot policy (i.e., the learned model) to estimate the value of different forms of real-time human feedback.
The main contributions of this work include 1) a method for estimating the value of different assistance mechanisms based on entropy measures in the robot action space and 2) evidence through a simulated study and user-study evaluation that the proposed method can estimate and select appropriate human interactions given the robot's uncertainty. In the remaining sections, we describe prior research in real-time interaction mechanisms, we describe our method for estimating the value of assistance mechanisms, we discuss the results of our evaluation, and we conclude with a general discussion related to our method for human-robot collaboration.
\section{Related Work}
\textbf{Mechanisms for Human Interaction---}
Past work has proposed several input mechanisms for human interaction with robot autonomy. Generally, these are characterized by the resulting level of autonomy (LOA) \cite{beer2014toward} of the robot in the human-robot team. Within the LOA, many methods involving real-time human interaction would generally be considered instances of shared control \cite{selvaggio2021autonomy}. For example; previous work has enabled high autonomy (e.g., supervisory) interfaces where humans provide discrete or preference feedback, real-time corrections or control over subspaces of the robot state space, and traded control where a human temporarily teleoperates at uncertain task moments. Often, the mediation of control handoffs is dictated by a measure over task/robot (e.g., action variance)  or human uncertainty \cite{dass2023pato,hoque2021thriftydagger, celemin2023knowledge, javdani2015shared, zurek2021situational, saeidi2015trust}.

\begin{figure*}[t]
\centering
\pdftooltip{\includegraphics[width=\textwidth]{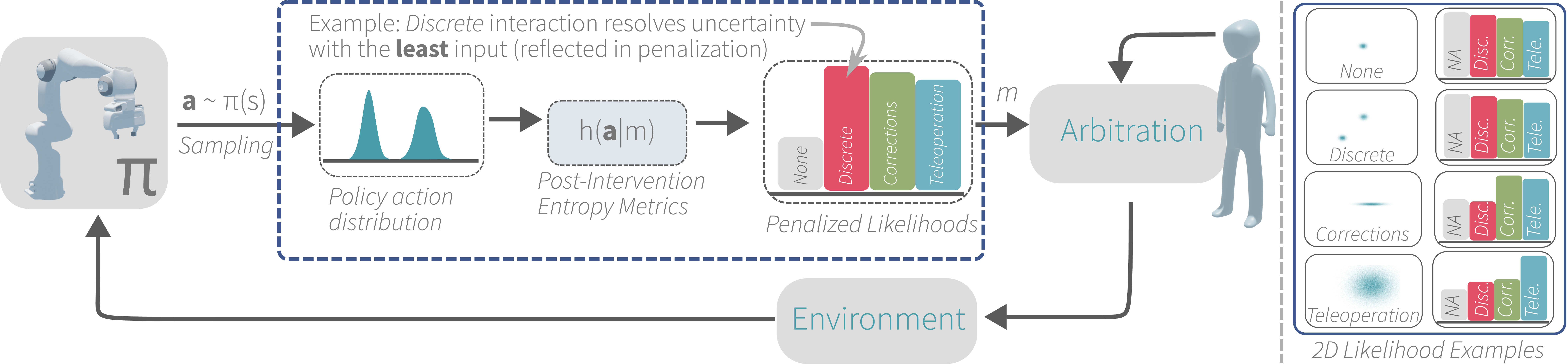}
\vspace{-15pt}}{The left part of the figure shows a block diagram with a 1D example where samples from the robot are used to build a action distribution from the robot. These feed into per-assistance-mechanism entropy measures which produces a bar plot of likelihoods for each mechanism. The highest likelihood is selected, which leads to a certain type of human input before the process repeats. The right shows 2D plots of action distributions and illustrative bar plots for likelihoods. At a high level, no uncertainty favors no assistance. A few modes in the 2D space favors discrete input. A line in the 2D space favors corrections. Finally, a wide uniform distribution favors teleoperation.}
\caption{Overview of the proposed method. \textit{Left:} We generate rollouts of a stochastic robot policy and calculate post-intervention differential entropy estimates to quantify the value of human assistance mechanisms ($m$) through a penalized likelihood. \textit{Right:} We provide illustrative 2D examples of how different action distributions map to different assistance likelihoods (the actual likelihoods are computed over a higher dimensional action space and over a time horizon).}
\label{fig:method_overview}
\vspace{-15pt}
\end{figure*}

\textbf{Approaches with Multiple Mechanisms---}
Several past works have developed methods that fuse multiple mechanisms for human input, but focus on robot task learning (i.e., refining a reward function) rather than effective real-time interaction. Often, these methods are categorized as interactive task learning or reinforcement learning from human feedback (RLHF) \cite{christiano2017deep}. At the most basic level, many methods allow the human to choose the most appropriate feedback to aid in robot learning \cite{blanchet2020guide,mehta2024unified,biyik2022learning, bullard2018towards}. Other methods combine multiple mechanisms under fixed interaction sequences (e.g., demonstrations followed by preferences) \cite{palan2019learning,ibarz2018reward}. To accommodate varied sources of human input, recent formalisms try to unify input types and their effects on learning \cite{jeon2020reward} and propose 
interfaces to learn from diverse human input \cite{biyik2022aprel, fitzgerald2023inquire}. Most relevant to our proposed method is INQUIRE \cite{fitzgerald2023inquire}, which uses information gain to reason about what interaction (e.g., demonstrations, corrections, or preferences) would lead to the largest information gain over uncertain robot reward parameters. Instead of interactive learning, our work focuses on real-time human interventions. For greater robot policy compatibility, we don't assume access to a reward function and instead use policy uncertainty in the robot's action space to guide user interventions. The coupling between robot actions and human input also necessitates a new formulation for mapping user interaction to expected entropy reductions.

\section{Estimating the Value of Assistance}
Our approach involves estimating the value of different human assistance mechanisms based on the real-time uncertainty of a robot policy (as illustrated in Figure \ref{fig:method_overview}). In this section, we describe the problem setting, our estimation method that leverages differential entropy and penalization, and finally practical considerations for implementing our approach.

\subsection{Problem Setting and Assumptions}
Similar to existing robot-gated assistance methods, our method assumes that policy variability is generally meaningful, requires human intervention, and that the human knows what needs to be done and can generally provide the correct input when asked (i.e., resolving the robot's uncertainty in a near-optimal manner). We find that policy variability often may signal a need for human intervention, such as when policy variability arises from a lack of data or from variable action data (e.g., different forces during surface finishing depending on surface characteristics) in similar robot states. While it is true that this assumption does not hold in all cases, such as where an imitation policy's variability is due to training on multiple valid trajectories or generally when variability doesn't impact task success, we still advocate that our input-minimizing method would be beneficial in potential false-positive cases (i.e., requesting assistance when unnecessary). Section \ref{sec:discussion} discusses mitigation strategies related to false-positive policy uncertainty and our human input assumptions.

Our approach is instantiated with a predefined set of human interaction mechanisms, $\mathcal{M}$. To estimate the mechanism value, we assume access to a learned robot policy with a stochastic and continuous action space, 
$\pi:\mathcal{S} \times \mathcal{A} \rightarrow [ 0, 1 ]$ (e.g., a diffusion model). Specifically, we leverage $n_r$ samples from the policy action distribution of a finite discrete-time horizon, $T_r$, from the current state, $\textbf{s}_\tau$ to estimate the assistance type values. We define tensor $\textbf{A}_\tau$ of the rollout actions at time $\tau$.
\begin{equation}
\begin{gathered}
    \textbf{A}_\tau \in \mathbb{R}^{n_a \times n_r \times T_r}:  \forall (n, t) \in  [0, ..., n_r-1] \times [0, ...,T_r-1], \\{[\textbf{A}_\tau]}_{:,n,t} \sim \pi(\textbf{s}_{n,t})
\end{gathered}
\end{equation}
where $n_a$ is the dimension of the action space and $\textbf{s}_{n,t}$ is the forecasted state. The horizon is necessary (as opposed to just the current action) to uncover behaviors such as bimodal junctions that diverge slowly in a continuous distribution. The finite horizon can be achieved through either a dynamics model of the policy or an action representation that explicitly includes a horizon (e.g., as is common in robot diffusion models \cite{chi2023diffusion}). It is assumed that the actions are absolute (e.g., pose instead of velocity) to allow for comparisons across the different actions at each time step in the forecasted behavior. In the case where the action space is differential, the differential actions can simply be aggregated over time. Ultimately, our goal is to define and estimate the value, $V(m \mid \pi, \textbf{s}_\tau) \doteq V(m \mid \textbf{A}_\tau)$, for each mechanism (i.e., $\forall m \in \mathcal{M}$), which is a trade-off between reduction in uncertainty and required human input. We drop $\pi$ and $\textbf{s}_\tau$ as they are implicitly represented in the rollouts, $\textbf{A}_\tau$.
Our problem setting dictates that the value of assistance mechanisms can be estimated online to propose real-time human interventions when the robot needs help. Thus, the sampling must be fast enough to use in the robot's decision-making loop.

\subsection{Value Estimation Approach}
\label{sec:approach}
Our premise is to compute assistance value estimates based on the policy action distribution. Our value estimates are based on characterizing the action uncertainty through an entropy formulation, as is common in the active learning literature  \cite{lopes2009active}. We next describe the specific mechanisms and parameterizations used in this work, followed by explaining our formulation to estimate the value of the assistance mechanisms. We end by discussing implementation considerations of our approach.

\subsubsection{Choice of Mechanisms}
There is a range of interaction types that have been considered previously in the robotics literature that vary by granularity of input \cite{koppol2021interaction}. In our work, we consider three common types: \textit{discrete preferences}, \textit{teleoperation}, and \textit{real-time corrections}. We also define a fourth estimate, \textit{no assistance}, where the robot operates without human input. Generally, our method is extensible to any interaction type where we can formulate an entropy-based value estimate and quantify the level of human input. As we are interested in continuous robot control, for each interaction, we leverage the post-intervention differential entropy as a measure of uncertainty.
\begin{equation}
h(\mathcal{A}_t|m,\pi) = -\int \limits_{\textbf{a} \in \mathcal{A}_t} p(\textbf{a} \mid m,\pi) \ln(p(\textbf{a} \mid m,\pi))
\end{equation}
where $\mathcal{A}_t$ is a random variable over the set of all possible actions at a given time step. Given that we have a forecast of robot actions, we will compute the differential entropy for each time step in horizon $T_r$. To simplify notation, we define $\textbf{A}_t \equiv {[\textbf{A}_\tau]}_{:,:,t}$ as the sample of actions for a given time step in the forecast. As the probability distribution of $\mathcal{A}_t$ is unknown, we will approximate the differential entropy by formulating expressions related to the policy rollouts, $\textbf{A}_\tau$ (i.e., $h(\textbf{A}_t|m)$). 

For each input mechanism ($m$), our method requires a definition for the expected post-intervention (i.e., the remaining uncertainty after human input) differential entropy ($h(\textbf{A}_t|m)$) as well as a measure of the human input, defined as $k_{m}$, that is used later as a penalization factor. We note that the formulation could equivalently be formulated through information gain because the pre-intervention entropy is the same across mechanisms. We define $k_{m}$ as the total input over the time horizon to differentiate mechanisms that require constant versus one-time input. Going forward, we describe the mechanisms in an order that simplifies notation.

\textbf{No Assistance--} The no user assistance/intervention ($m_\varnothing$) is for when the robot can act confidently without the aid of the human operator. In this case, the value of post-intervention entropy is simply the policy uncertainty, and our estimate is the sample entropy from the policy's action distribution:
\begin{equation}
h(\textbf{A}_t \mid m_{\varnothing}) \coloneq \hat{h}(\textbf{A}_t)
\end{equation}
where $\hat{h}$ is the sample entropy. This mechanism requires no operator input (i.e., $k_{m_{\varnothing}}\coloneq 0$).

\textbf{Discrete Input--} The discrete mechanism ($m_{d(n_d)}$) is when the operator is presented with a finite set of choices for what behavior the robot executes. We define our estimate for a fixed number of discrete choices (e.g., 2 versus 3 choices) and the system can maintain and choose between different value estimates for the different number of discrete choices (with different penalizations). For a discrete estimate with $n_d$ choices, we then cluster the rollouts (using random restarts K-means clustering over the entire forecasted horizon) and compute the expected entropy over the clusters. Intuitively, this clustering approach should only significantly reduce the entropy if the data falls nicely into $n_d$ clusters, such as when the rollouts consist of distinct paths.

\begin{equation}
    \mathcal{C} = \argmin\limits_{\mathcal{C}} \sum\limits_{i}^{n_d} \sum\limits_{n \in \mathcal{C}_i}  || [\textbf{A}_\tau]_{:,n,:}-\textbf{M}_i ||_{F}^{2}
\end{equation}
\begin{equation}
h(\textbf{A}_t \mid m_{d(n_d)}) \coloneq  \sum\limits_{i}^{n_d} \frac{|\mathcal{C}_i|}{n_r}\hat{h}({[\textbf{A}_\tau]}_{:,n \in \mathcal{C}_i,t})
\end{equation}
where $\mathcal{C}$ is the set of clusters (of size $n_d$), $||\cdot||$ is the Frobenius norm, $\textbf{M}_i \in \mathbb{R}^{n_a \times T_r}$ are the cluster means, and $|\mathcal{C}_i|$ is the number of trajectories in a cluster. As an example, if the forecasted robot behavior contains two distinct paths (e.g., toward one of two objects), the weighted-sum differential entropy over the two separated behaviors will be much lower than the differential entropy of the full forecast. If there are not clear clusters (e.g., a uniform distribution) or one of the clusters has high entropy, the weighted-sum entropy over clusters will still be relatively high, suggesting the discrete interaction might not be appropriate. The required user input is the selection of a discrete path. We represent the input as the number of choices (i.e., $k_{m_{d(n_d)}} \coloneq n_d$) to differentiate the increased effort of selecting from larger discrete options.

\textbf{Teleoperation--} The teleoperation mechanism ($m_r$ for \textit{remote}) is where the operator temporarily controls the robot's actions when the robot's behavior is highly uncertain of the correct action to execute. We assume that the operator is able to completely resolve the policy uncertainty (i.e., they know what action to perform) and executes the action in a noisily optimal way (e.g., due to interface error in providing teleoperation input \cite{aker2012assessing}). We can accordingly model the probability of the operator selecting an action as:

\begin{equation}
p(\textbf{a}_t \mid m_r,\pi) \propto e ^{ -\beta||  \mathbf{a}_t-\boldsymbol{\zeta}_t|| }
\end{equation}
where $\beta$ is the level of optimality and $\boldsymbol{\zeta}_t$ is the desired optimal human action. With the assumptions of noisily optimal action selection and that the error in the human's input is independent across variables (i.e., a diagonal covariance matrix) and with the same variance (derived from $\beta$), the differential entropy can be expressed in closed form \cite{cover1999elements} as:
\begin{equation}
\label{eq:gaussianent}
h(\textbf{A}_t \mid m_{r}) \coloneq \frac{1}{2} \ln  (\beta^{-n_a}) + \frac{n_a}{2} \bigl(1+ \ln(2\pi)\bigr)
\end{equation}
The modeling assumptions can also be removed if the human error covariance can be directly estimated. Note that this estimate is a constant that only depends on the dimension of the action space and the estimated noise of the teleoperator (i.e., it is independent of $\textbf{A}_t$). The input required from the human is the product of the size of the action space and length of the horizon (i.e., $k_{m_r} \coloneq n_a \cdot T_r$).

\textbf{Differential Corrections--} The correction mechanism ($m_{c(n_c)}$) is when the operator provides differential corrections to a subspace of the robot control. For example, an operator might provide precise adjustments to the rotation of a peg as it is inserted into a hole or adjustments to the applied force during surface finishing operations. Similar to the discrete mechanism, we define the estimate for a fixed number of operator control dimensions ($n_c$) and can maintain separate estimates for different correction dimensions. Our approach to differential corrections closely follows the formulation of Hagenow et al. \cite{hagenow2021informing} where the singular value decomposition (SVD) and principal component analysis (PCA) are used to extract out the most likely corrections at each time step. Intuitively, the entropy estimate here is the sample entropy after projecting noisily optimal human actions onto the first $n_c$ principal directions (i.e., the maximum variance subspace of the actions) that the operator will control and leaving the original policy actions for the remaining robot-controlled for the remainder of the action basis.
\begin{equation}
\begin{gathered}
\textbf{U}\boldsymbol{\Sigma}\textbf{V} = \textrm{SVD}(\textbf{A}_t-\bar{\textbf{A}}_t) \\
\textbf{A}^{'}_t = [\textbf{V}^{\textrm{T}}_{0...n_c} \; \textbf{0} ]^{\textrm{T}}\textbf{A}^{\textrm{C}} + [\textbf{0} \; \textbf{V}^{\textrm{T}}_{n_c...n_a} ]^{\textrm{T}}(\textbf{A}_t-\bar{\textbf{A}}_t) \\
\forall a_c \in \textbf{A}^{\textrm{C}}, a_c \sim \mathcal{N}(0,1/\beta) \\
h(\textbf{A}_t \mid m_{c(n)}) \coloneq \hat{h}(\textbf{A}^{'}_t)
\end{gathered}
\end{equation}
where $\bar{\textbf{A}}_t$ is the mean forecast trajectory, $\textbf{U}\boldsymbol{\Sigma}\textbf{V}$ is the SVD (vectors and values), $\textbf{A}^{\textrm{C}}$ is a matrix of Gaussian samples, and $\textbf{A}^{'}_t$ is a superposition of samples from the human model Gaussian and the original robot actions (the composition of which is determined by the correction dimension). For implementation modularity, we draw samples from a Gaussian in the correction directions and compute the sample entropy. Alternatively, the estimate could combine the analytical entropy for the correction axes and empirical entropy for the remaining axes. 
 The input required from the human is the product of the number of correction dimensions and length of the horizon (i.e., $k_{m_{c(n_c)}} \coloneq n_c \cdot T_r, n_c < n_a$). The benefit of the PCA approach is that the correction directions (i.e., principal components) can be efficiently calculated from the forecasted robot actions. We are interested in exploring nonlinear methods in future work.


\subsubsection{Value Estimate of Assistance}
To determine the most valuable human assistance mechanism, we construct a penalized likelihood for each candidate mechanism based on the post-intervention expected action uncertainty and the complexity of the mechanism. First, for each mechanism, we use the post-intervention entropy to define a mechanism likelihood. When the post-intervention outcome is certain, the differential entropy will be low and when the policy is uncertain, the differential entropy will be high. Unlike the Shannon entropy (i.e., the discrete formulation), the differential entropy is unbounded. While there are many options to convert the entropy to a likelihood, we leverage upper and lower entropy bounds.

\begin{equation}
V(m \mid \textbf{A}_\tau) \coloneq  \frac{ \lambda_m\left(T_r h_{\textrm{max}}-\sum \limits_{t}^{T_r} h(\textbf{A}_{t} \mid m)\right)}{T_r (h_{\textrm{max}}-h_{\textrm{min}})}
\label{eq:norm}
\end{equation}
where $\lambda_{m}$ is the mechanism penalization factor and $h_{\textrm{max}}$ and $h_{\textrm{min}}$ are the maximum and minimum entropy respectively. We assume the human's noisily optimal actions represent the minimum uncertainty and thus, use Equation \ref{eq:gaussianent} for $h_{\textrm{min}}$. For the maximum uncertainty, we define a conservative entropy estimate using a uniform distribution over the range of action values from the policy model (e.g., in our implementation, the minimums and maximums of the training data):
\begin{equation}
h_{\textrm{max}} \coloneq \text{log} \left( \prod \limits_{i}^{n_a} (\textbf{a}_{i,\textrm{max}} - \textbf{a}_{i,\textrm{min}}) \right)
\label{eq:hmax}
\end{equation}
where $\textbf{a}_{i,\textrm{max}}$ and $\textbf{a}_{i,\textrm{min}}$ are the maximum and minimum possible action values respectively. 
When such range values cannot be easily obtained or estimated, Equations \ref{eq:norm} and \ref{eq:hmax} can be trivially replaced with a different normalizing function (e.g., sigmoid).
In establishing the mechanism penalization factors (i.e., $\lambda_m$), we penalize the likelihood based on the expected human input in the intervention mechanism. Specifically, each mechanism requires $k_m$ input, and for equal likelihoods, we would prefer the mechanism that minimizes human input. Thus, we constrain the penalization factors accordingly.
\begin{equation}
\label{eq:penalization}
    \forall (m_i,m_j)\in \mathcal{M}\times\mathcal{M},  k_{m_i} > k_{m_j} \implies \lambda_{m_i}<\lambda_{m_j}
\end{equation}
We discuss practical considerations and strategies for setting the penalization parameters in the next section. With the value estimation defined for each mechanism; at each step in the robot behavior, we can determine the best-suited interaction by selecting the highest-value mechanism.
\begin{equation}
m^{*} = \argmax_{m \in \mathcal{M}} V(m \mid \textbf{A}_\tau)
\end{equation}

\pagebreak
\subsection{Practical Considerations for Implementation}
\label{sec:methodpractical}
Implementing our approach requires important design decisions, such as the choice of policy and penalization. Here we describe key implementation details. We also provide open-source code for the estimation algorithm and our experimental environments (including data generation) on Github \footnote{\href{https://github.com/mhagenow01/REALM_algorithm}{https://github.com/mhagenow01/REALM\_algorithm}}.

\textbf{Policy and Behavior Sampling--}
We desire to sample from the robot's action distribution for real-time assistance inference. For all of our experiments, we use an adapted version of Chi et al.'s Diffusion Policy \cite{chi2023diffusion}, which uses a state-conditioned Denoising Diffusion Probabilistic Model (DDPM) with Denoising Diffusion Implicit Models (DDIM) to improve the speed of inference. We chose DDPM for its ability to capture various action distributions (e.g., multi-modal behaviors). Model details are in the experimental section.

\textbf{Sample Entropy Estimates --} In our implementation, we compute the sample entropy ($\hat{h}$) based on the estimator from Ebrahimi et al. \cite{ebrahimi1994two}. We lower bound the estimates based on Equation \ref{eq:gaussianent} (i.e., the noisily optimal human input is treated as the lowest achievable entropy). Empirically, the sample entropy estimate often underestimates the true entropy in particular action variables (as opposed to the sum over action variables) and thus, we lower bound each action variable (i.e., $h_i = max(h_i,-\frac{1}{2} \ln  (\beta) + \frac{1}{2} \bigl(1+ \ln(2\pi)\bigr))$.


\textbf{Penalization Parameters --}
While Equation \ref{eq:penalization} suggests a straightforward hierarchy of assistance penalization factors, determining the penalization values can be challenging; as the appropriate value is a function of the action size, the number of samples in the entropy calculation, and the range of the action variables. We describe our tuning in the next section. Practically, we advocate for starting with a small amount of penalization (favoring higher levels of control under uncertainty) and refining the parameters from there. We also imagine a future iterative process where the weights are refined from post-hoc user feedback (further enabling personalization).


\begin{figure*}[]
\centering
\pdftooltip{\includegraphics[width=\textwidth]{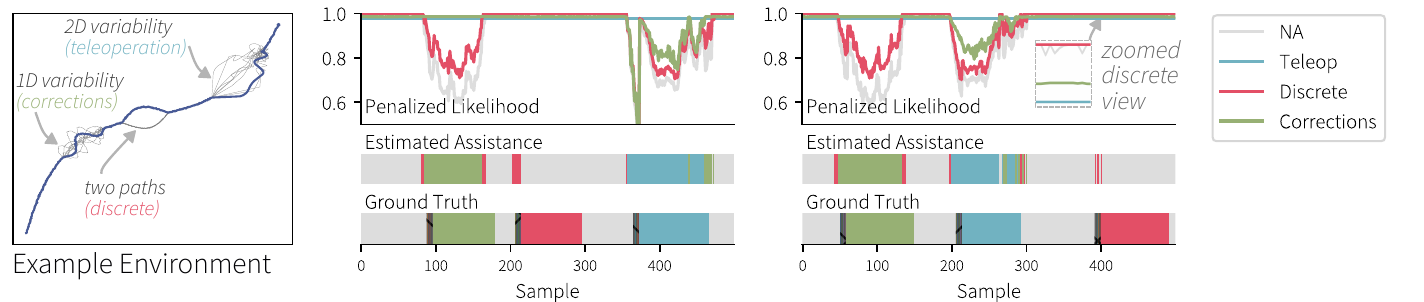}}{Three plots of the simulated environment. The left one shows an example of the 2D simulation and uncertainty which includes a two-way junction, as well as gray samples of the 1D and 2D path uncertainty for corrections and teleoperation, respectively. The Right two plots show examples of the likelihoods, estimates, and ground truth for \method. Generally, these two plots show good agreement between ground truth and the estimates though there are a few missed classifications near the edges of the teleoperation and correction behaviors.}
\vspace{-20pt}
\caption{Example results from \textit{Uncerpentine}. \textit{Left:} Example of a randomly generated environment (blue indicates the test path [i.e., ground truth] and gray are a subset of the trajectory data). \textit{Right:} Examples (the left is from the same environment) of the penalized likelihoods, ground truth, and assistance estimation. Gray hatch indicates the margin before assistance within the behavior forecast horizon. The right example shows policy collapse in teleoperation.}
\label{fig:uncerpentine}
\vspace{-15pt}
\end{figure*}



\section{Experimental Results}
To assess and demonstrate the value of our proposed approach, we conducted two experiments. The first experiment was a simulated 2D navigation task and the purpose was to \emph{assess the accuracy and tendencies of our mechanism estimation approach} (compared to a known ground truth policy uncertainty). The second experiment was a preliminary user study consisting of an uncertain tabletop manipulation task in which participants compared our multi-mechanism approach to a common uncertainty-based shared control approach. The purpose of the second experiment was to \emph{assess the impact of our approach on a representative task in terms of task performance, time, and perceived usability}.

\subsection{Learning Model and Training Details}
Both the simulated experiment and robot user study leveraged state-conditioned diffusion models for the robot policy. The only model difference was the training data (i.e., action space and number of training trajectories). The diffusion model parameters are shown in Table \ref{table:diffusion_details}. All model training was performed using an NVIDIA Volta V100. For the robot experiment, the model was deployed to an NVIDIA GeForce RTX 3070 Ti. With this GPU, we were able to achieve a control frequency of 7-8 Hz during the evaluation.

\begin{table}[h]
    \centering
    \begin{tabular}{@{}ll@{}}
        \toprule
        \textbf{Training Details} &  \\
        \midrule
        Number of epochs & 100 \\
        Batch size & 2068 (\textit{sim}) / 1577 (\textit{robot}) \\
        Training time & 127h (\textit{sim}) / 9.5h (\textit{robot}) \\
        Learning rate & 1e-4 \\
        Optimizer & AdamW \\
        Loss function & MSE \\
        Diffusion Steps (DDIM) & 10 \\
        Noise Schedule & Squared Cosine \\
        Network & Conditional UNet \\
        \bottomrule
    \end{tabular}
    \vspace{10pt}
    \caption{Diffusion Model Training and Architecture Details}
    \label{table:diffusion_details}
    \vspace{-20pt}
\end{table}

\subsection{Simulated Experiment}
\textbf{Task and Metrics--} We developed a 2D navigation environment, \textit{Uncerpentine}, that consisted of randomly generated Bézier curves with injected sinusoidal variability (e.g., 2D, 1D, and discrete junctions to match with the desirable assistance types) at randomized time steps. An example is shown in Figure \ref{fig:uncerpentine}. We compared our method to the ground-truth variability type (and corresponding assistance type). Here the ground-truth refers to the type of injected uncertainty (in the training data) at each time step in the test trajectory. In this experiment, there is no user control. For each time step, We evaluate the state in the test trajectories to see whether the system selects the appropriate human interaction based on the known uncertainty in the training data. For discrete interactions, as the variability collapses once a path is chosen, rather than comparing to ground truth, we search for whether the system estimates a discrete interaction prior to the junction.

\textbf{Training and Testing Data --}
We generated 10 environments (two for setting penalization weights and eight for evaluation). For each environment, we generated 1000 sample trajectories. 900 samples were used to train the environment diffusion model and 100 samples were used for testing (i.e., evaluating the estimated assistance against the known training variability as a ground truth). Each trajectory had a total length of 600 time steps. The first 500 time steps were used for testing. The final 100 time steps repeated the same final action to encourage the Diffusion Policy to converge to a final state. When running the assistance estimation, we generated 50 rollout samples with a diffusion prediction horizon of 64 time steps. Only the first 16 time steps are used to calculate the value of assistance through the entropy estimates (i.e., $T_r=16$). This longer diffusion model horizon was empirically necessary for the policy to better capture the data uncertainty. 

\textbf{Results and Discussion--} Table \ref{table:confusion} provides the confusion matrix results and qualitative examples appear in Figure \ref{fig:uncerpentine}. For \textit{Discrete} interactions, we had a recognition rate of $1.0$ (i.e., no missed discrete interactions). Generally, our approach was able to recognize the different interactions. Many of the misclassifications occurred in transition. Due to the finite horizon prediction that acts as a filter, we did not expect to see correct classification near the boundaries, which accounted for approximately 15 percent of the samples.
\begin{table}[b]
    \centering
    \vspace{-10pt}
    \begin{tabular}{@{}cc cccc@{}}
    \multicolumn{1}{c}{} &\multicolumn{1}{c}{} &\multicolumn{2}{c}{Estimated} \\ 
    \cmidrule(lr){3-6}
    \multicolumn{1}{c}{} & 
    \multicolumn{1}{c}{Actual} & 
    \multicolumn{1}{c}{N.A.} & 
    \multicolumn{1}{c}{Corrections} & 
    \multicolumn{1}{c}{Teleoperation} & 
    \multicolumn{1}{c}{Discrete} \\ 
    \cline{2-6}
    & N.A.  & \textbf{0.856} & 0.045 & 0.049 & 0.05   \\
    & Corrections  & 0.110   & \textbf{0.805} & 0.003 & 0.081 \\ 
    & Teleoperation & 0.004 & 0.118 & \textbf{0.870} & 0.008 \\
    \cline{2-6}
    \end{tabular}
    \vspace{10pt}
    \caption{\textit{Uncerpentine} Confusion Matrix}
     \label{table:confusion}
    \vspace{-15pt}
\end{table}
The most common misclassification was between corrections and teleoperation. In our 2D environment, there was less separation (i.e., 1D vs 2D input) than would be expected in robot control (e.g., 6 DOF). Post-hoc inspection showed that many of the other misses were caused by policy mode collapse (i.e., not capturing the full data uncertainty). We believe that estimate filtering strategies to aid in switching and policy under-approximations of the uncertainty will be useful in future implementations.


\begin{figure*}[]
\centering
\pdftooltip{\includegraphics[width=\textwidth]{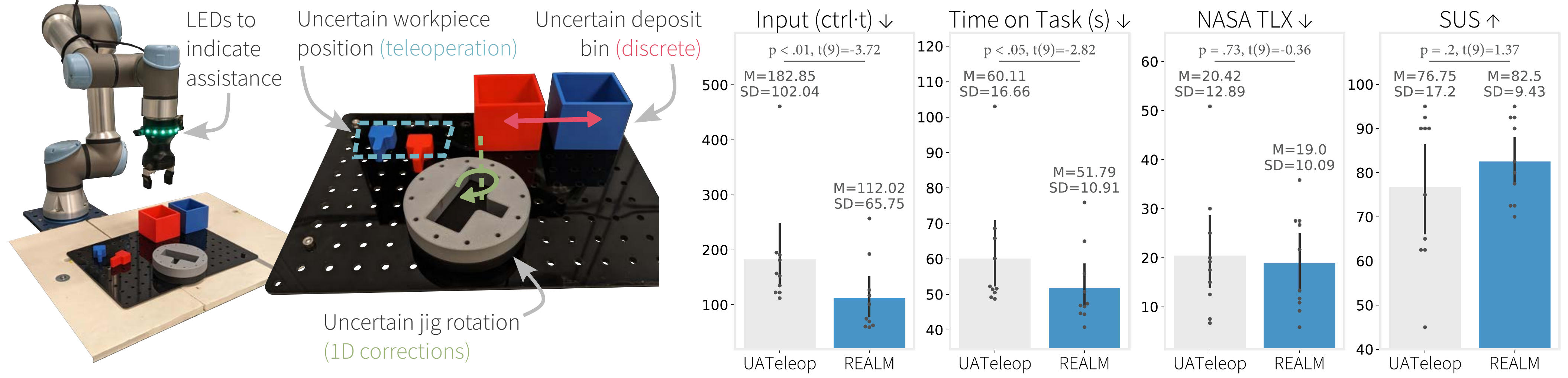}}{The left pane shows the task and robot setup with a zoomed view of the task. This zoomed view highlights the sources of uncertainty for each task step. The right plot shows four bar plots with statistics for the study metrics.}
\vspace{-20pt}
\caption{\textit{Left}: Study task with three assistance regions. \textit{Right}: Study results and statistics.}
\label{fig:userstudy}
\vspace{-10pt}
\end{figure*}

\subsection{User Evaluation of Robot Manipulation Task}
\label{sec:userstudy}
We conducted a preliminary study with 10 participants (5M, 5F), aged 20--47 ($M=27.9, SD=7.2$), with little-to-moderate robot familiarity ($2$ \emph{not familiar}, $3$ \emph{slightly}, $5$ \emph{moderately}), recruited from campus lists under exempt protocol (E-5797) from the MIT Institutional Review Board (IRB). Our exploratory study was designed to assess interactions with \method compared to a common shared control approach.  

\textbf{Task and Metrics--} The task was an abstracted machine tending task with injected variability in the training data (see Figure \ref{fig:userstudy}). The task consisted of three main steps (also shown in the supplemental video). First, the robot reaches down and grasps a key-shaped workpiece. Second, the robot raises the workpiece above the gray jig and then lowers and rotates into the jig to fit the shape. Finally, the robot raises the piece out of the jig and deposits it in a red or blue bin depending on the color of the workpiece. This task was inspired by uncertain machine tending tasks. Here the uncertainty comes from the robot being trained without vision data (e.g., only the person can see where the piece is, how the jig is rotated, and the color of the bin where it should be deposited), though we imagine applying the same approach to visuomotor policies.

In the experiment, participants practiced with each condition three times prior to a recorded trial. The order of the conditions was counterbalanced. The different trials had randomized sources of uncertainty (four starting locations for the workpiece, three rotations for the jig, and the color of the workpiece). After each recorded trial, participants filled out the NASA TLX \cite{hart1988development} and Systems and Usability Scale (SUS) \cite{bangor2008empirical} questionnaires. We also recorded the total task time and amount of user input. The input was based on the assistance mode and control dimension (measured in $\textrm{control-timesteps})$:
\begin{equation}
    input = 0\cdot t(NA)+1\cdot t(Corr)+5\cdot t(Tele)+|Discrete|
\end{equation}
where $t(\cdot)$ refers to the time in a mode and $|Discrete|$ refers to the number of discrete decisions. The five in teleoperation accounts for control of three Cartesian directions, one rotation, and the gripper. Results were analyzed using a paired T-test. We also collected participant feedback and preferences.

\textbf{Conditions--} \textit{REALM (ours):}  The three interactions (corrections, teleoperation, and discrete) were provided by participants using a SpaceMouse (e.g., to control the robot and provide discrete feedback with the buttons). LEDs on the robot end effector indicated the requested assistance. In our implementation, we also included action chunking (i.e., computing new forecasts, but executing behaviors from the same forecast for several steps) and requiring multiple consecutive teleoperation estimates (to reduce errant switching) before engaging in teleoperation, given the high handover cost.

\textit{Uncertainty-Aware Teleoperation (UATeleop):} We modeled our baseline after robot-gated approaches that leverage robot action variance to elicit human teleoperation \cite{menda2019ensembledagger,celemin2023knowledge,dass2023pato}. Using a single-intervention baseline also allowed us to quantify \method's reduction of user input and to assess if the minimized human input affects task success compared to common teleoperation takeover approaches. With our robot policy, we mediate hand-offs between the robot and operator teleoperation by thresholding the robot's action variance.
\begin{equation}
\textbf{a} = 
\begin{cases}
    [\textbf{A}_\tau]_{:,0,t} & \sum \limits_{t}^{T_r} \textrm{var}((\textbf{A}_t)) \leq \gamma \\
    \textbf{a}_h & else
\end{cases}
\end{equation}
where $\textbf{a}_h$ is the human action and $\gamma$ is a tuned threshold that determines when teleoperation is needed. While we also desired a multi-mechanism comparison, existing methods required reward functions and thus were incompatible with and not easily modified for our robot imitation policy.

For both \method and \textit{UATeleop}, we empirically tuned the penalization parameters to provide the best user experience (i.e., making sure the operator could provide assistance when the robot needed help and trying to minimize errant assistance requests). Empirically, we found that only a small number of iterations (fewer than 10) of parameter tuning (following the hierarchy of Equation \ref{eq:penalization}) were necessary to achieve satisfactory behavior. This led to the following penalization parameters. In \method, we used $\lambda = [1.0, 0.862, 0.954, 0.885]$ (the order corresponded to \textit{no assist}, \textit{teleoperation}, \textit{discrete}, \textit{corrections}). Our testing indicated that ranges of penalization values that would produce the same assistance patterns (e.g., $\geq2\%$ for any parameter) and that larger parameter perturbations tended to still lead to task success, but with unnecessary interaction requests. For \textit{UATeleop}, we set the switching threshold to $\gamma=0.3$. We found difficulty selecting a penalization parameter that reliably recognized the discrete junction without creating false-positive teleoperation.

\textbf{Policy Training Data--}
We created 1500 robot manipulation trajectories to train our user study diffusion policy in a Robosuite \cite{zhu2020robosuite} simulated environment of our task where each iteration had a randomized workpiece starting location, jig rotation, and workpiece color. Each trajectory was 300 time steps long, though consistent trajectory length is not a required. Similar to the simulated experiment, the last 70 time steps repeated the same final action to encourage the Diffusion Policy to converge to a final state. The state consisted of the robot end-effector position, orientation, gripper state, and a Boolean for whether the gripper had entered the jig. The action space consisted of the desired robot end-effector position, orientation, and gripper state. When running the assistance estimation during the user study, we generated 50 samples with a prediction horizon of 64 time steps  and an estimate horizon of 12 time steps. The action chunking was for 8 time steps if there was no change in assistance. We found that a longer diffusion prediction horizon helped avoid hallucination (e.g., sampling more than two actions at the bin junction).

\textbf{User Interactions and Signaling--}
In our manipulation experiment, participants engaged in three assistance modes indicated by LED colors near the robot gripper of assistance. White LEDs indicated no assistance was required. Future versions may consider human-initiated takeovers if the robot is mistakenly confident. Green LEDs indicated the participant could teleoperate the robot in 4-DOF (position and yaw rotation) and toggle the gripper using the SpaceMouse. We removed the other two rotation axes that were unnecessary for the task and caused participant mistakes during pilots. Participants could hand back control to the robot using a button when it flashed, indicating robot confidence. Yellow LEDs indicated corrective assistance where participants controlled the first principal component of variability, which tended to rotate the end effector (during jig insertion). We cached the state and correction direction at the beginning of corrective behaviors to avoid continuity issues. We have plans for more general correction handling in future work. Red LEDs indicated discrete input was needed and prompted the user to select between two paths. Participants could not see visualized paths in the study and were told the buttons corresponded to the left and right options. After selection, the robot executed the full horizon of actions to ensure it passed the discrete junction. 

\textbf{Results and Discussion--} Figure \ref{fig:userstudy} shows the survey results and statistical results from our study. All participants completed the task successfully using both conditions. Two participants repeated trials due to user errors. One error was in \method condition where the participant triggered a programmatic (conservative) safety stop for permissible motion. The second error occurred during \textit{UATeleop} where the participant mistakenly and unrecoverably dropped the workpiece.

We found that \method required significantly less time and input. While the input reduction was expected, we were encouraged that the minimized input also led to more time-efficient interactions in our study. \method did not significantly increase task load (there were also no significant differences [$p<0.05$] for any of the TLX subscales) or decrease usability as might be expected by the addition of user interaction types. This might suggest that the targeted and minimized elicitation of input from REALM balances the overhead of managing several interaction types. Seven out of 10 participants indicated that they preferred \method over \textit{UATeleop}. Those who preferred \method commented that it reduced input and the robot appeared more knowledgeable. Those who preferred \textit{UATeleop} preferred the simplicity of one mode and the greater control.
\section{General Discussion}
\label{sec:discussion}
We presented, \textit{REALM}, a method for estimating the most valuable human assistance to a robot based on entropy estimates of a stochastic robot policy's actions. We showed how these estimates can be computed accurately in real time and enable input-efficient user interactions.

\textbf{Limitations and Future Work--} We assume an uncertain robot policy and that policy uncertainty needs human intervention. Future methods would benefit from human feedback of when requests are unnecessary to adapt and avoid future repeat requests. While we did not focus on learning, if the robot model can be retrained or fine-tuned, it is possible that the human's assistive actions can improve model performance \cite{christiano2017deep}. Our current method assumes a noisily-optimal human model with a fixed $\beta$ and also that the human can provide near-optimal input. In the future, we are interested in modeling the $\beta$ parameter from human data \cite{ghosal2023effect} and  exploring different human models.
Our evaluation contained only synthetic data. Future work will investigate existing datasets, models (including how often current models would request assistance), and Learning from Demonstration. We only compared REALM to one common shared control approach. Future studies will compare different approaches to multi-input design (e.g., methods to tune penalization, different personalization approaches, different interface designs). Our simulated study highlighted practical failure situations (e.g., policy mode collapse) that we will address in future work. Our penalization factors were tuned for empirical performance (rather than an automated or iterative approach) and not systematically assessed for parameter sensitivity. We only evaluated single hypotheses over discrete and corrective inputs (i.e., two paths and 1D corrections). Regarding human-robot teaming, a practical instantiation of our method requires investigating design choices related to human factors (e.g., input and communication mechanisms).





\bibliographystyle{IEEEtran}
\bibliography{references}

\begin{thebibliography}{10}
\providecommand{\url}[1]{#1}
\csname url@samestyle\endcsname
\providecommand{\newblock}{\relax}
\providecommand{\bibinfo}[2]{#2}
\providecommand{\BIBentrySTDinterwordspacing}{\spaceskip=0pt\relax}
\providecommand{\BIBentryALTinterwordstretchfactor}{4}
\providecommand{\BIBentryALTinterwordspacing}{\spaceskip=\fontdimen2\font plus
\BIBentryALTinterwordstretchfactor\fontdimen3\font minus \fontdimen4\font\relax}
\providecommand{\BIBforeignlanguage}[2]{{%
\expandafter\ifx\csname l@#1\endcsname\relax
\typeout{** WARNING: IEEEtran.bst: No hyphenation pattern has been}%
\typeout{** loaded for the language `#1'. Using the pattern for}%
\typeout{** the default language instead.}%
\else
\language=\csname l@#1\endcsname
\fi
#2}}
\providecommand{\BIBdecl}{\relax}
\BIBdecl

\bibitem{chi2023diffusion}
C.~Chi, Z.~Xu, S.~Feng, E.~Cousineau, Y.~Du, B.~Burchfiel, R.~Tedrake, and S.~Song, ``Diffusion policy: Visuomotor policy learning via action diffusion,'' \emph{The International Journal of Robotics Research}, 2023.

\bibitem{yoneda2023noise}
T.~Yoneda, L.~Sun, G.~Yang, B.~C. Stadie, and M.~R. Walter, ``To the noise and back: Diffusion for shared autonomy,'' in \emph{Robotics: Science and Systems XIX, Daegu, Republic of Korea, July 10-14, 2023}, 2023.

\bibitem{ng2023diffusion}
E.~Ng, Z.~Liu, and M.~Kennedy, ``Diffusion co-policy for synergistic human-robot collaborative tasks,'' \emph{IEEE Robotics and Automation Letters}, 2023.

\bibitem{beer2014toward}
J.~M. Beer, A.~D. Fisk, and W.~A. Rogers, ``Toward a framework for levels of robot autonomy in human-robot interaction,'' \emph{Journal of human-robot interaction}, vol.~3, no.~2, p.~74, 2014.

\bibitem{selvaggio2021autonomy}
M.~Selvaggio, M.~Cognetti, S.~Nikolaidis, S.~Ivaldi, and B.~Siciliano, ``Autonomy in physical human-robot interaction: A brief survey,'' \emph{IEEE Robotics and Automation Letters}, vol.~6, no.~4, pp. 7989--7996, 2021.

\bibitem{dass2023pato}
S.~Dass, K.~Pertsch, H.~Zhang, Y.~Lee, J.~J. Lim, and S.~Nikolaidis, ``Pato: Policy assisted teleoperation for scalable robot data collection,'' in \emph{Robotics: Science and Systems}, 2023.

\bibitem{hoque2021thriftydagger}
R.~Hoque, A.~Balakrishna, E.~Novoseller, A.~Wilcox, D.~S. Brown, and K.~Goldberg, ``Thriftydagger: Budget-aware novelty and risk gating for interactive imitation learning,'' in \emph{5th Annual Conference on Robot Learning}.

\bibitem{celemin2023knowledge}
C.~Celemin and J.~Kober, ``Knowledge-and ambiguity-aware robot learning from corrective and evaluative feedback,'' \emph{Neural Computing and Applications}, vol.~35, no.~23, pp. 16\,821--16\,839, 2023.

\bibitem{javdani2015shared}
S.~Javdani, S.~S. Srinivasa, and J.~A. Bagnell, ``Shared autonomy via hindsight optimization,'' \emph{Robotics science and systems: online proceedings}, vol. 2015, 2015.

\bibitem{zurek2021situational}
M.~Zurek, A.~Bobu, D.~S. Brown, and A.~D. Dragan, ``Situational confidence assistance for lifelong shared autonomy,'' in \emph{2021 IEEE International Conference on Robotics and Automation (ICRA)}.\hskip 1em plus 0.5em minus 0.4em\relax IEEE, 2021, pp. 2783--2789.

\bibitem{saeidi2015trust}
H.~Saeidi and Y.~Wang, ``Trust and self-confidence based autonomy allocation for robotic systems,'' in \emph{2015 54th IEEE Conference on Decision and Control (CDC)}.\hskip 1em plus 0.5em minus 0.4em\relax IEEE, 2015, pp. 6052--6057.

\bibitem{christiano2017deep}
P.~F. Christiano, J.~Leike, T.~Brown, M.~Martic, S.~Legg, and D.~Amodei, ``Deep reinforcement learning from human preferences,'' \emph{Advances in neural information processing systems}, vol.~30, 2017.

\bibitem{blanchet2020guide}
K.~Blanchet, A.~Bouzeghoub, S.~Kchir, and O.~Lebec, ``How to guide humans towards skills improvement in physical human-robot collaboration using reinforcement learning?'' in \emph{2020 IEEE International Conference on Systems, Man, and Cybernetics (SMC)}.\hskip 1em plus 0.5em minus 0.4em\relax IEEE, 2020, pp. 4281--4287.

\bibitem{mehta2024unified}
S.~A. Mehta and D.~P. Losey, ``Unified learning from demonstrations, corrections, and preferences during physical human--robot interaction,'' \emph{ACM Transactions on Human-Robot Interaction}, vol.~13, no.~3, pp. 1--25, 2024.

\bibitem{biyik2022learning}
E.~B{\i}y{\i}k, D.~P. Losey, M.~Palan, N.~C. Landolfi, G.~Shevchuk, and D.~Sadigh, ``Learning reward functions from diverse sources of human feedback: Optimally integrating demonstrations and preferences,'' \emph{The International Journal of Robotics Research}, vol.~41, no.~1, 2022.

\bibitem{bullard2018towards}
K.~Bullard, A.~L. Thomaz, and S.~Chernova, ``Towards intelligent arbitration of diverse active learning queries,'' in \emph{2018 IEEE/RSJ International Conference on Intelligent Robots and Systems (IROS)}.\hskip 1em plus 0.5em minus 0.4em\relax IEEE, 2018, pp. 6049--6056.

\bibitem{palan2019learning}
M.~Palan, G.~Shevchuk, N.~Charles~Landolfi, and D.~Sadigh, ``Learning reward functions by integrating human demonstrations and preferences,'' in \emph{Robotics: Science and Systems}, 2019.

\bibitem{ibarz2018reward}
B.~Ibarz, J.~Leike, T.~Pohlen, G.~Irving, S.~Legg, and D.~Amodei, ``Reward learning from human preferences and demonstrations in atari,'' \emph{Advances in neural information processing systems}, vol.~31, 2018.

\bibitem{jeon2020reward}
H.~J. Jeon, S.~Milli, and A.~Dragan, ``Reward-rational (implicit) choice: A unifying formalism for reward learning,'' \emph{Advances in Neural Information Processing Systems}, vol.~33, pp. 4415--4426, 2020.

\bibitem{biyik2022aprel}
E.~B{\i}y{\i}k, A.~Talati, and D.~Sadigh, ``Aprel: A library for active preference-based reward learning algorithms,'' in \emph{2022 17th ACM/IEEE International Conference on Human-Robot Interaction (HRI)}.\hskip 1em plus 0.5em minus 0.4em\relax IEEE, 2022, pp. 613--617.

\bibitem{fitzgerald2023inquire}
T.~Fitzgerald, P.~Koppol, P.~Callaghan, R.~Q. J.~H. Wong, R.~Simmons, O.~Kroemer, and H.~Admoni, ``Inquire: Interactive querying for user-aware informative reasoning,'' in \emph{Conference on Robot Learning}.\hskip 1em plus 0.5em minus 0.4em\relax PMLR, 2023, pp. 2241--2250.

\bibitem{lopes2009active}
M.~Lopes, F.~Melo, and L.~Montesano, ``Active learning for reward estimation in inverse reinforcement learning,'' in \emph{Joint European Conference on Machine Learning and Knowledge Discovery in Databases}.\hskip 1em plus 0.5em minus 0.4em\relax Springer, 2009, pp. 31--46.

\bibitem{koppol2021interaction}
P.~Koppol, H.~Admoni, and R.~G. Simmons, ``Interaction considerations in learning from humans.'' in \emph{IJCAI}, 2021, pp. 283--291.

\bibitem{aker2012assessing}
A.~Aker, M.~El-Haj, M.-D. Albakour, U.~Kruschwitz \emph{et~al.}, ``Assessing crowdsourcing quality through objective tasks.'' in \emph{LREC}, 2012, pp. 1456--1461.

\bibitem{cover1999elements}
T.~Cover, \emph{Elements of information theory}.\hskip 1em plus 0.5em minus 0.4em\relax John Wiley \& Sons, 1999.

\bibitem{hagenow2021informing}
M.~Hagenow, E.~Senft, R.~Radwin, M.~Gleicher, B.~Mutlu, and M.~Zinn, ``Informing real-time corrections in corrective shared autonomy through expert demonstrations,'' \emph{IEEE Robotics and Automation Letters}, vol.~6, no.~4, pp. 6442--6449, 2021.

\bibitem{ebrahimi1994two}
N.~Ebrahimi, K.~Pflughoeft, and E.~S. Soofi, ``Two measures of sample entropy,'' \emph{Statistics \& Probability Letters}, vol.~20, no.~3, pp. 225--234, 1994.

\bibitem{hart1988development}
S.~G. Hart and L.~E. Staveland, ``Development of nasa-tlx (task load index): Results of empirical and theoretical research,'' in \emph{Advances in psychology}.\hskip 1em plus 0.5em minus 0.4em\relax Elsevier, 1988, vol.~52, pp. 139--183.

\bibitem{bangor2008empirical}
A.~Bangor, P.~T. Kortum, and J.~T. Miller, ``An empirical evaluation of the system usability scale,'' \emph{Intl. Journal of Human--Computer Interaction}, vol.~24, no.~6, pp. 574--594, 2008.

\bibitem{menda2019ensembledagger}
K.~Menda, K.~Driggs-Campbell, and M.~J. Kochenderfer, ``Ensembledagger: A bayesian approach to safe imitation learning,'' in \emph{2019 IEEE/RSJ International Conference on Intelligent Robots and Systems (IROS)}.\hskip 1em plus 0.5em minus 0.4em\relax IEEE, 2019, pp. 5041--5048.

\bibitem{zhu2020robosuite}
Y.~Zhu, J.~Wong, A.~Mandlekar, R.~Mart{\'\i}n-Mart{\'\i}n, A.~Joshi, S.~Nasiriany, and Y.~Zhu, ``robosuite: A modular simulation framework and benchmark for robot learning,'' \emph{arXiv preprint arXiv:2009.12293}, 2020.

\bibitem{ghosal2023effect}
G.~R. Ghosal, M.~Zurek, D.~S. Brown, and A.~D. Dragan, ``The effect of modeling human rationality level on learning rewards from multiple feedback types,'' in \emph{Proceedings of the AAAI Conference on Artificial Intelligence}, vol.~37, no.~5, 2023, pp. 5983--5992.

\end{thebibliography}


\end{document}